\documentclass[letterpaper, 10 pt, conference]{ieeeconf}

\newcommand{\papertitle}{Real-time Coupled Centroidal Motion and Footstep Planning for Biped Robots}

\usepackage{graphics} 
\usepackage{epsfig} 
\usepackage{mathptmx} 
\usepackage{times} 
\usepackage{amsmath} 
\usepackage{amssymb}  
\usepackage{bm}

\usepackage{float} 
\usepackage{color} 
\usepackage[hidelinks]{hyperref} 
\hypersetup{breaklinks=true}

\usepackage{algorithm}
\usepackage{algpseudocode}
\usepackage{optidef}
\usepackage{mdframed}
\usepackage{tocloft}

\usepackage{etoolbox}
\apptocmd{\sloppy}{\hbadness 10000\relax}{}{}
\apptocmd{\sloppy}{\vbadness 10000\relax}{}{}

\usepackage{soul}

\title{\LARGE \bf
\papertitle
}

\usepackage{supertabular,array}             
\usepackage{booktabs}                       
\usepackage{tabularx}                       

\usepackage{textcomp}                       
\usepackage{gensymb}                        
\usepackage[printonlyused,nolist]{acronym}  

\usepackage{svg}
\usepackage{multirow}

\makeatletter
\renewcommand*{\ALG@name}{Problem}
\makeatother

\author{Tara Bartlett and Ian R. Manchester%
\thanks{The authors are with the Australian Centre for Robotics (ACFR),
and the School of Aerospace, Mechanical and Mechatronic Engineering, The University of Sydney, Sydney, NSW 2006, Australia.
       {\tt\small tara.bartlett@sydney.edu.au}
}
}

\begin{document}






\begin{acronym}[\hspace{\NomenLHSwidth}\hspace{1em}]
    \newcommand{\comm}[1]{\acroextra{ \emph{(#1)}}}
    \setlength{\itemsep}{0pt}
    \setlength{\parskip}{0pt}

    %
    \acro{USyd}{University of Sydney}
    \acro{ACFR}{Australian Centre for Robotics}
    \acro{NASA}{National Aeronautical and Space Administration}
    \acro{JPL}{the NASA Jet Propulsion Laboratory}
    \medskip
    \acro{FSP}{Footstep Planning}
    \acro{TO}{Trajectory Optimisation}
    \acro{OSC}{Operational Space Controller}
    \acro{MPC}{Model Predictive Control}
    \acro{CLF}{Control Lyapunov Function}
    \acro{SLIP-MPC}{SLIP MPC} 
    \medskip
    \acro{DoF}{Degrees of Freedom}
    \acro{CG}{Center of Gravity}
    \acro{CoM}{Center of Mass}
    \acro{SLAM}{Simultaneous localisation and mapping}
    \acro{lidar}{light detection and ranging sensor}
    \acro{IMU}{inertial measurement unit}
    \acro{KF}{Kalman filter}
    \acro{EKF}{extended \acl*{KF}}
    \acro{PD}{Proportional-Derivative}
    \acro{VO}{Visual Odometry}
    \acro{VIO}{Visual Inertial Odometry}
    \medskip
    \acro{SSP}{Single Support Phase}
    \acro{DSP}{Double Support Phase}
    \acro{FP}{Flight Phase}
    \medskip
    \acro{ZMP}{Zero-Moment Point}
    \acro{HZD}{Hybrid Zero Dynamics}
    \medskip
    \acro{LIP}{Linear Inverted Pendulum}
    \acro{SLIP}{Spring-Loaded Inverted Pendulum}
    \acro{ALIP}{Angular Momentum \acl*{LIP}}
    \acro{TMIP}{Two-Mass Inverted Pendulum}
    \acro{MMIP}{Multiple-Masses Inverted Pendulum}
    \acro{VHIP}{Virtual Height Inverted Pendulum}
    \medskip
    \acro{QP}{Quadratic Program}
    \acro{MIP}{Mixed Integer Program}
    \acro{QCQP-SDR}{Quadratically-Constrained Quadratic Program with Semi-Definite Relaxation}
    \acro{ADMM}{Alternating Directions Method of Multipliers}
    \acro{ISTA}{Iterative Shrinkage and Threshold Algorithm}
    \acro{FISTA}{Fast Iterative Shrinkage and Threshold Algorithm}
    \acro{KKT}{Karush-Kahn-Tucker}
    \medskip
    \acro{KNN}{K Nearest Neighbours}
    \acro{RKNN}{RK Nearest Neighbours}
    \medskip
    \acro{IRIS}{Iterative Regional Inflation by Semidefinite programming}
    \acro{DEM}{Digital Elevation Map}
    \acro{RGB}{Red-Green-Blue}
    \acro{RGBD}{Red-Green-Blue-Depth}
    \acro{PCD}{Pointcloud}
    \medskip
    \acro{MSR}{Mars Sample Return}
    \acro{SRL}{Sample Retrieval Lander}
    \acro{MAV}{Mars Ascent Vehicle}
    \acro{SRH}{Sample Recovery Helicopter}
    \acro{NEPA}{National Environmental Policy Act}
    \medskip
    \medskip
    \acro{TTL}{Tansistor-Transistor Logic}
    \acro{ROS}{Robot Operating System}
    \acro{ROS2}{Robot Operating System 2}
    \acro{MoCap}{Motion Capture}
    \medskip
    \acro{CFA}{Cumulative Fractional Area}
    \acro{GRC-3}{Martian soil simulant Glenn Research Center 3}
    \acro{ABIT*}{Advanced Batch Informed Trees}
    \acro{ARA*}{Anytime Repairing A*}
    \acro{RGG}{Random Geometric Graph}
    \acro{LPA*}{Truncated Lifelong Planning A*}
    \acro{OMPL}{Open Motion Planning Library}
    \medskip
    \acro{NN}{Neural Network}
    \acro{DNN}{Deep Neural Network}
    \acro{CNN}{Convolutional Neural Network}
    \acro{RNN}{Recurrent Neural Network}
    \acro{RL}{Reinforcement Learning}
    \acro{FL}{Fuzzy Logic}
    \acro{GA}{Genetic Algorithms}
    \medskip
    \acro{SOTA}{State of the Art}
    \medskip
    \acro{MuJoCo}{Multi-Joint dynamics with Contact simulator}
\end{acronym}


\maketitle
\thispagestyle{empty}
\pagestyle{empty}


\begin{abstract}
This paper presents an algorithm that finds a centroidal motion and footstep plan for a \acf{SLIP}-like bipedal robot model substantially faster than real-time. This is achieved with a novel representation of the dynamic footstep planning problem, where each point in the environment is considered a potential foothold that can apply a force to the center of mass to keep it on a desired trajectory.
For a biped, up to two such footholds per time step must be selected, and we approximate this cardinality constraint with an iteratively reweighted $l_1$-norm minimization. Along with a linearizing approximation of an angular momentum constraint, this results in a quadratic  program can be solved for a contact schedule and center of mass trajectory with automatic gait discovery. A 2~s planning horizon with 13 time steps and 20 surfaces available at each time is solved in 142~ms, roughly ten times faster than comparable existing methods in the literature. We demonstrate the versatility of this program in a variety of simulated environments.
\end{abstract}


\section{Introduction}\label{sec:intro}
Motion planning for legged robotic systems involves both finding a \ac{CoM} trajectory and a set of footsteps. With a discrete set of steps defining the continuous CoM trajectory, it is a tightly coupled problem that is difficult to solve when considering both the dynamics of the \ac{CoM} and variation in the terrain.
Existing real-time methods simplify the problem by limiting the plans to statically stable motion \cite{Tonneau2020} or assuming consistency in the terrain \cite{Huang2021, Gibson2022}, limiting the flexibility of the system. Other methods are solved in a duration similar to or longer than the length of the horizon \cite{winkler_gait_2018, Ponton2021} which impacts the feasibility of real-time application with low-powered on-board computing.

In this paper, we consider the centroid of the robot as a point mass, where the trajectory is determined by the cumulative sum of the accelerations resulting from forces applied over time. We start by supposing that all nearby surfaces can apply a force to the \ac{CoM} to keep it close to a desired trajectory, and then find a sparse subset of such forces to attain feasibility with a limited number of legs.
In particular, we represent the bipedal configuration as a point-mass body with two spring-loaded, telescopic, massless legs as per a \ac{SLIP} model. This model constrains the reaction forces to be parallel to the leg vector (from the foothold to the \ac{CoM} position) and allowing up to two contact forces at any given time for double support phases.

\begin{figure}[tbp]
    \centering
    \includegraphics[width=\linewidth]{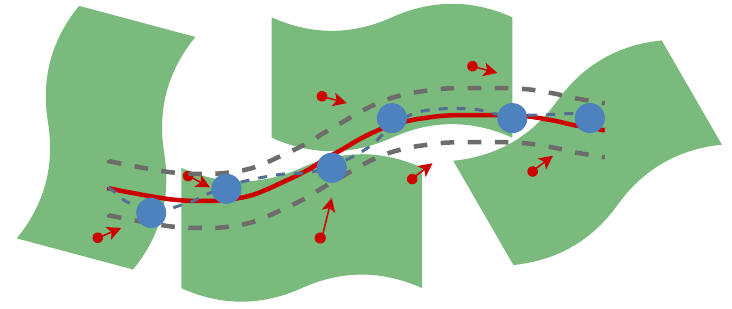}
    \caption{Conceptual depiction of the cumulative sum of forces to follow the desired \ac{CoM} trajectory: the green patches represent traversable terrain;  the red line is the desired path bounded by the gray dashed lines; the blue dot and dashed line represent the \ac{CoM} trajectory; and the red arrows are the forces applied.}
    \label{fig:rtdt-research-concept}
\end{figure}

The key concept is to consider the steps as forces applied to the robot by the contact points in the environment, rather than the feet pushing on the contact points. The acceleration of the \ac{CoM} is defined by the sum of the forces applied by each step location such that a sparse combination of feasible steps (and forces to apply) can be selected to follow a desired trajectory.This is depicted in Figure \ref{fig:rtdt-research-concept}.

\begin{figure}[tbp]
    \centering
    \includegraphics[width=1\linewidth]{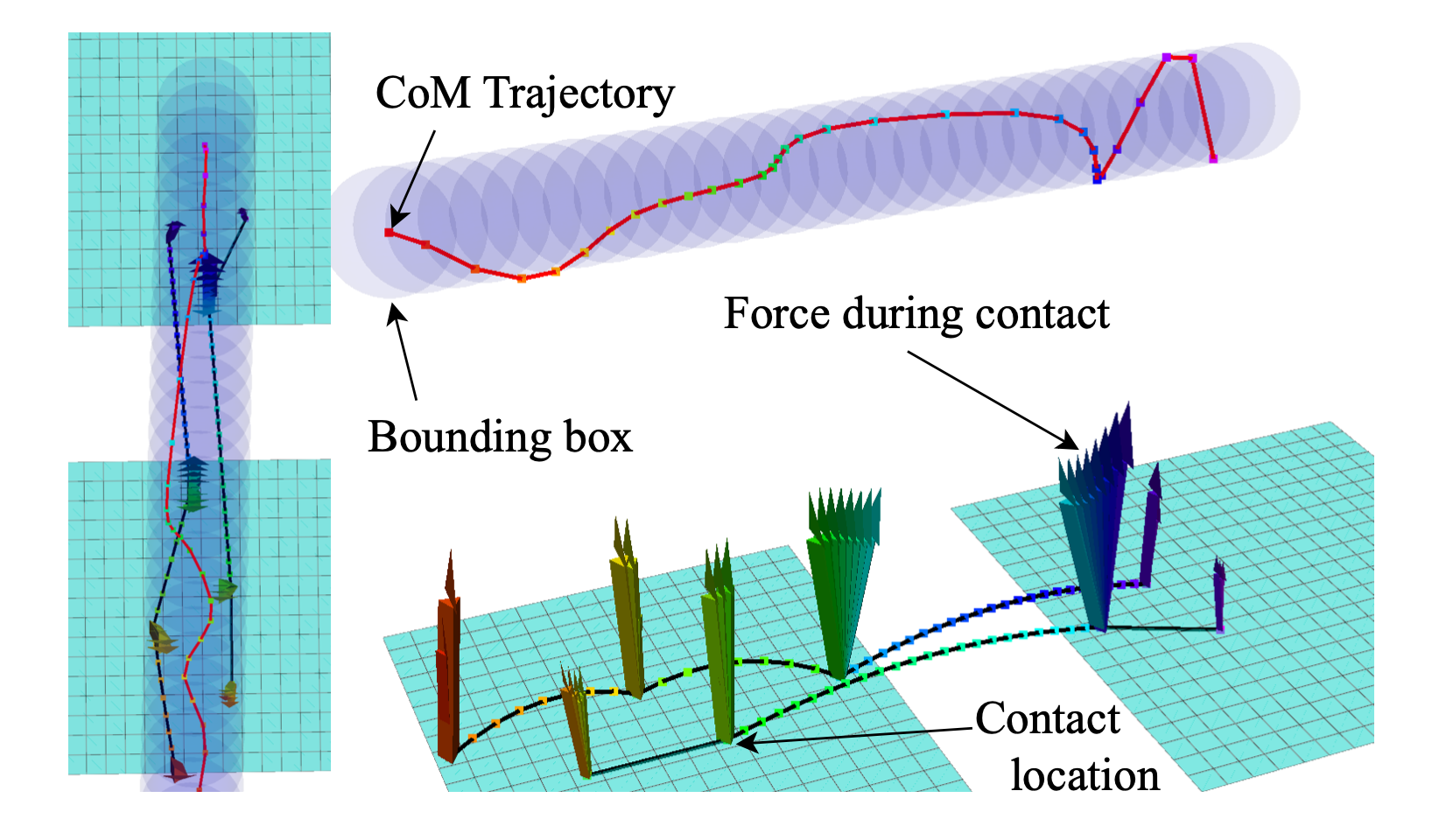}
        \caption{Example trajectory computed by the proposed program. Given a terrain cost map and a high-level desired trajectory, a \ac{CoM} trajectory and footstep plan is generated for a \acf{SLIP}-like robot model jumping over a chasm. The 4.8s horizon is solved in 0.544~s with 20 available contact locations at each time step.}
    \label{fig:cover}
\end{figure}

We present a convex approximation of this problem which generates solutions like Fig. \ref{fig:cover}, providing a coupled \ac{CoM} and footstep plan. This is achieved significantly faster than real-time, in the sense that the solve time on an Intel Core i7 CPU/3.00GHz 8-Core desktop is substantially less than the planning time horizon. This algorithm assumes availability of a traversability cost map (e.g. a grid map \cite{Fankhauser2016}) and a global path or heading from a higher level planner or remote operator. The generated plan and contact schedule can also provide a warm start for higher fidelity trajectory optimization algorithms for bipedal robots \cite{apgar2018fast, Xiong2021} in receding horizon frameworks.

\begin{figure*}[tbp!]
    \centering
    \includegraphics[width=\linewidth]{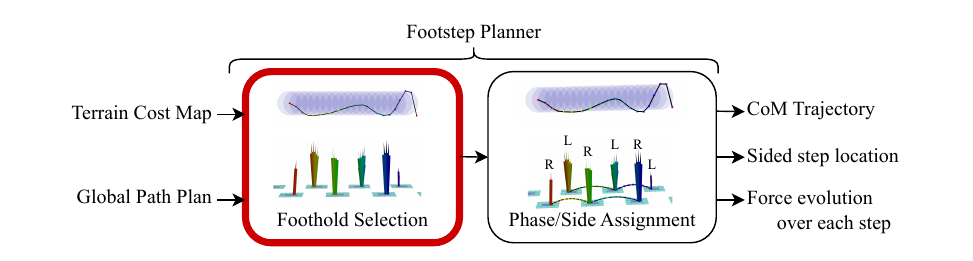} 
    \caption{Methodology overview: foothold selector calculates a set of forces to apply, phase assignment groups the forces into contact phases and assigns a side to the steps. The output goes to a trajectory optimizer that calculates \ac{CoM} and foot trajectories, which can then enter an operational space controller \cite{apgar2018fast} to calculate motor torques.}
    \label{fig:system-overview}
\end{figure*}

\subsection{Related Work}\label{sec:literature}
This section provides an overview of online motion planning methods for bipedal robots that do not perform pre-calculations, such creating as a library of pre-defined gaits to be selected from. These are either limited to statically-stable motion, do not consider the terrain, or do not explicitly solve the footstep planning problem (assuming consistency in terrain characteristics). 

\subsubsection{Real-time bipedal footstep planning for \textbf{statically-stable} motion with terrain consideration}
Due to the complexity of the problem, body motion is often constrained or assumed to be statically stable \cite{griffin2019footstep, Tonneau2020}, allowing for the dynamics and traversability problems to remain decoupled, or simplified, such that the solutions can be calculated in real-time. Acosta \cite{acosta2023bipedal} presents a program that can plan dynamic footstep plans but is constrained to walking motion.

\subsubsection{Real-time bipedal footstep planning for dynamic motion on \textbf{flat ground}}
Recent work has involved the development of more dynamic trajectories \cite{gong2019cassie, li2019using}, allowing for more time and energy efficient motion but do not consider the terrain. Gibson \cite{Gibson2022} accounts for the terrain in the foot placement calculation but assumes consistent terrain characteristics (piece-wise planar terrain) to determine how to place the steps to optimize the dynamics of the robot.

\subsubsection{Real-time bipedal \textbf{motion} planning for dynamic motion with terrain consideration}
Huang \cite{Huang2021} uses a potential field representation in a reactive planner encouraging the robot towards the goal but does not explicitly solve the footstep planning problem, assuming some continuity or holistic structure in the environment, as opposed to selecting footholds to allow for inconsistencies or infeasible regions in the terrain.

\subsubsection{Real-time \textbf{quadrupedal} footstep planning for dynamic motion with terrain consideration}
Robots with four or more legs are easier to stabilize due to the increased support polygon. This has been well progressed for quadrupedal robots solving the footstep planning problem in real-time with predictive control with whole-body dynamics \cite{Mastalli2022, bjelonic2021whole}.

\subsubsection{\textbf{Offline} bipedal footstep planning for dynamic motion with terrain consideration}
Other methods have involved selecting footstep locations in a map of the environment then matching that to a pre-calculated library of gaits \cite{Kuindersma2020}, which require the step transitions and contact points to be comparable to those included in the data used to learn the gait library. Methods with longer solve times employ complementarity constraints \cite{dafarra2022dynamic}, bilevel optimization structures \cite{zhu_contact-implicit_2021}, mixed-integer programs \cite{Ponton2021}, or phase-based end-effector parametrization \cite{winkler_gait_2018}.

\subsection{Contributions and Paper Structure}

This paper contributes:
\begin{itemize}
    \item A  computationally efficient coupled \ac{CoM} motion and footstep planning program for dynamic motion of bipedal robots with consideration of the terrain; and
    \item Validation with a variety of simulated test environments and path configurations.
\end{itemize}

The problem is defined in Section \ref{sec:prob-def} and the convex approximation presented in Section \ref{sec:quad-prog} is validated with the trials shown in Section \ref{sec:results}.

\section{Optimization Program Formulation}\label{sec:prob-def}
The robot model considered is a \ac{SLIP}-like model, where the robot is defined by a point mass at the \ac{CoM} with two spring-loaded, telescopic, massless legs.

\subsection{Problem Inputs}\label{sec:inputs}
The set of feasible steps available at each time step is based on the traversability cost map, $C$, and foot reach based on the desired \ac{CoM} position at the corresponding sample time. At each time, there are $K$ potential contact locations identified within a reachable radius, $R$, of the \ac{CoM} position using a \ac{RKNN} search of surfaces, $e\in \mathbb{E}$, in a provided map of the environment.  The range of available forces to apply at each surface, $e$, is constrained by the friction cone, the magnitude of force the robot can apply, and ensuring the forces are parallel to the leg vector (constraining the centroidal angular momentum to zero), as in a SLIP model. This is expressed in Problem \ref{alg:nl-nc-fsp}, the elements of which are detailed in this section.

\subsection{Optimization Variables}\label{sec:variables}
Variables of the optimisation program are in bold. The trajectory is discretized with resolution $t$ with $N$ sample times. The variables are $\mathbf{s}$, an $N\times 3$ matrix of 3D \ac{CoM} positions at $N$ sample times, and $\mathbf{a}$, an $N\times K\times 3$ matrix containing 3D acceleration vectors in cartesian coordinates for each available contact location at each time.

\subsection{The Constraints}\label{sec:constraints}

\subsubsection{Representation of the environment} \label{sec:feasible-steps}
As per Fig. \ref{fig:system-overview}, it is assumed that a global path/heading and a terrain map of surfaces with associated costs, surface normals, and friction coefficients are available. A set of reachable surfaces is selected via a \ac{RKNN} search. This is provided as input to the algorithm through the environment matrix, $C$. The corresponding accelerations of the \ac{CoM} are accumulated to define the trajectory of the \ac{CoM} over time.



\subsubsection{Relating footsteps to CoM motion}\label{sec:kinematics}
Discretizing the trajectory into time steps with a constant sample time, $t$, the kinematic relationship is expressed as
\begin{align}
\begin{bmatrix}
	\mathbf{s}_o\\
	\vdots\\
	\mathbf{s}_N
\end{bmatrix}_{[N \times 3]}
= t^2
\begin{bmatrix}
	\frac{1}{2}&&0\\
	\vdots&\ddots&\\
	\frac{2N-1}{2}&\dots&\frac{1}{2}
\end{bmatrix}_{[N \times N]}
\sum\limits_{i=0}^K\left(\mathbf{a} - g \right)  + s_0\label{eqn:kinematics-matrix}
\end{align}
where $s_0$ is the initial position of the CoM and $g$ is the gravity vector. This is expressed as
\begin{align}
\mathbf{s}&= P\sum\limits_{i=0}^K\left(\mathbf{a} - g \right) + s_0.\label{eqn:kinematic-shortform}
\end{align}

This can be calculated along each axis separately. At each time, the position, $\mathbf{s}$, is constrained to within a tolerance of the desired path, $s^*$.

\subsubsection{Friction cone}
The friction cone determines the maximum tangential components, $\mathbf{a}_{t_1}$ and $\mathbf{a}_{t_2}$, that can be applied to the surface without slip, given the normal component, $\mathbf{a}_{n}$, of the resultant acceleration:
\begin{align}
    \mathbf{a}_n\mu \ge \sqrt{\mathbf{a}_{t_1}^2 + \mathbf{a}_{t_2}^2}. \label{eqn:friction-nonlin}
\end{align}

\subsubsection{Direction of the force} \label{sec:fsp-force-dir}
As the leg is represented as a spring in the SLIP model, the force must be towards the \ac{CoM}, parallel to the leg vector between the CoM positions and the contact locations, $f_p$, as per Equation \eqref{eqn:force-dir}. Thus, in a SLIP model, the angular momentum of the centroid is constrained to zero at each available contact location at each time:
\begin{align}
    \mathbf{a} \times (\mathbf{s} - f_p) = 0. \label{eqn:force-dir}
\end{align}

\subsubsection{Number of Contacts}
As the bipedal robot has two feet, there must be no more than two contact points at any given time, as per the cardinality constraint
\begin{align}
    \text{card}(\mathbf{a}_n) \le 2. \label{eqn:cardinality}
\end{align}
By constraining the number of contact points at any time, the program has the flexibility to determine the gait automatically without prior specification of the number of phases or phase durations.

\subsection{The Objective Function}
The objective contains encourages a better selection of footholds and a smooth CoM trajectory that closely follows the target.

\subsubsection{Selecting a sparse set of steps weighted by the terrain cost map}
To encourage the planner towards selecting single support phases for walking that are placed based on the terrain cost map, an $l_1$-norm cost term is used to encourage sparsity, weighted by the environment matrix cost map using the Hadamard product
\begin{align}
    f_{env} = \left|C\circ\mathbf{a}_n\right|_1.\label{eqn:cardinality-obj}
\end{align}

\subsubsection{Path following}
To follow the path, the $l_2$-norm of the deviation from the desired position and velocity is penalized with
\begin{align}
    f_{path} = \left|\mathbf{s}-\Tilde{s}\right|_2  + w\left|\frac{d\mathbf{s}}{dt}-\frac{d\mathbf{s^*}}{dt}\right|_2.
\end{align}

\subsubsection{Temporal consistency}
By encouraging preference for forces at consecutive sample times to be on the same surface (as opposed to on different, neighboring surfaces), a series of impulse-like steps is replaced by a longer-duration single stance phase. This leads to more consistent foot placement over time. The resulting \ac{CoM} motion is not notably different, but the foot trajectories are more representative of a typical walking gait.

This is achieved by adding an $l_2$-norm cost term on the difference in acceleration on each surface, $e$, in the environment map, $\mathbb{E}$, between consecutive time steps, as per Equation \eqref{eqn:temporal-consistency}. The cost is only applied to the normal component of the acceleration as the tangential components are then defined by the constraint on the force direction presented in Section \ref{sec:constraints}.
\begin{align}
    f_{tc}= \sum\limits_{e\in \mathbb{E}}\left|\frac{d\mathbf{a}_n}{dt}\right|_2. \label{eqn:temporal-consistency}
\end{align}

\subsubsection{Penalty on jerk}
To encourage smoothness in the \ac{CoM} motion, a cost term is added to penalize the jerk of the \ac{CoM}, $j$, the derivative of acceleration:
\begin{align}
    f_{jerk} = \left|j\right|_2.
\end{align}

\subsubsection{The complete objective function}
The presented terms are combined into the weighted convex quadratic objective function below,
\begin{align}
   obj =& w_0f_{env} + w_1f_{tc} + w_2f_{path} + w_3f_{jerk}    \\
   =& \min\limits_{\mathbf{s},\mathbf{a}} w_0\left|C\circ\mathbf{a}_n\right|_1 + w_1\sum\limits_x \left|\frac{d\mathbf{a}_n}{dt}\right|_2\\
   &+ w_2\left(\left|\mathbf{s}-\Tilde{s}\right|_2 + w\left|\frac{d\mathbf{s}}{dt}-\frac{d\mathbf{s^*}}{dt}\right|_2\right) + w_3\left|j\right|_2,
\end{align}

forming the foothold selector program in Problem \ref{alg:nl-nc-fsp}.


\begin{algorithm}[tbp]
    \caption{Coupled dynamic traversability bipedal motion and footstep selection problem.}
    \label{alg:nl-nc-fsp}
    \bgroup
    \def\arraystretch{1.3} 
    \hspace{-3mm}
    \begin{tabular}{clr}
    find        & $\mathbf{s}_{[i,3]} \in \mathbb{R}^3$    &  (\ac{CoM} position)\\
                & $\mathbf{a}_{[i,k,3]} \in \mathbb{R}^3$             & (step acceleration)\\
    $\min\limits_{\mathbf{s},\mathbf{a}}$ & $w_0\left|C\circ\mathbf{a}_n\right|_1+ w_1\sum\limits_x \left|\frac{d\mathbf{a}_n}{dt}\right|_2 $ &\\
                &$+ w_2\left(\left|\mathbf{s}-s^*\right|_2 + w\left|\frac{d\mathbf{s}}{dt}-\frac{d\mathbf{s^*}}{dt}\right|_2\right)$&(cost function) \\
                &$+ w_3\left|j\right|_2$& \\
    s.t. & $\mathbf{a}\le a_{max}$                                            &(max acceleration)\\
                & card$(\mathbf{a}_n) \le 2   $                    & (max two contacts)\\
                & $\left|\mathbf{s} - s^* \right| \le tol$                                         & (path tolerance) \\
                &  $ a_n \mu \ge \sqrt{a_{t_1}^2 + a_{t_2}^2}$             & (friction constraint) \\
                
                & $\mathbf{s}=s_0 + P\left(\mathbf{a} - g\right)$              & (kinematics) \\
                & $\mathbf{a} \times (\mathbf{s}-f_p)= 0$& (force direction)\\
    \end{tabular}
    \egroup
\end{algorithm}

\section{Convex Approximation}\label{sec:quad-prog} 
To improve the run time, the nonlinear, non-convex problem above is approximated by the quadratic program presented in Problem \ref{alg:convex-fsp}.

\subsection{Direction of the Force}
Equation \eqref{eqn:force-dir} is a nonlinear constraint, since both the acceleration, $\mathbf{a}$, and \ac{CoM} position, $\mathbf{s}$, are decision variables. This is approximated to the linear representation in Equation \eqref{eqn:force-dir-constraint}, using an approximation of the \ac{CoM} position, $\tilde{s}$, which is initially approximated as the desired state, $s^*$.
The extent of the inaccuracy is limited by the path tolerance constraint. 
The acceleration vector is then parallel to the approximation of the leg vector that is known ahead of time, such that
\begin{align}
    a = \bm{\alpha}\circ(\Tilde{s} - f_p)\label{eqn:force-dir-constraint}
\end{align}

where $\bm{\alpha}$ is a scale factor for each surface available at each time, defining a smaller set of decision variables and reducing the solve time.

\subsection{Friction cone}
With the approximated constraint on the force direction, the force that can be applied at a given surface at a given time is known before the problem is set up. This allows for acceleration vectors outside the friction cone to be eliminated from the set of feasible footholds prior to running the program.

\subsection{Weighted Sparsity by Cost Map}

The cardinality equation, denoting the weighted sparsity in the step selection (Equation \eqref{eqn:cardinality-obj}), is approximated by minimizing the $l_1$-norm of $\bm{\alpha}$. This is known to encourage sparsity by minimizing the cardinality (see \cite{candes_enhancing_2007} and references therein). Given that the normal component must be greater than or equal to zero, the sum of the absolute values is equivalent to the sum and therefore simplifies to 
\begin{align}
    \text{min} \{\text{card}(x)\} \approx & \text{ min} \left|x\right|_1 = \text{min}\sum\limits_{i=1}^N x\\
    f_{env}&= \sum\limits_{i=1}^NC\circ\bm{\alpha}. \label{eqn:fsp-obj-sparsity}
\end{align}

\begin{algorithm}[tbp]
    \caption{Convex approximation of Problem \ref{alg:nl-nc-fsp}.}
    \label{alg:convex-fsp}
    \bgroup
    \def\arraystretch{1.3} 
    \hspace{-3mm}
        \begin{tabular}{clr}
find        & $\bm{\alpha}_{[i,k]} \in \mathbb{R}$      &  (step scale factor)\\
    $\min\limits_{\bm{\alpha}}$ & $w_0\sum\limits_{i=1}^NC\circ\bm{\alpha} + w_1\sum\limits_{e\in \mathbb{E}} \left|\frac{d\bm{\alpha}}{dt}\right|_2 $ &\\
                &$+ w_2\left(\left|s-s^*\right|_2 + w\left|\frac{ds}{dt}-\frac{ds^*}{dt}\right|_2\right)$&(cost function) \\
                &$+ w_3\left|j\right|_2 + w_4 \frac{\left|\bm{\alpha}^r\right|}{\left|\bm{\alpha}^{r-1}\right| + \epsilon}$ & \\
    s.t.        & $a\le a_{max}$                                            &(max acceleration)\\
                & $\bm{\alpha} \ge 0 $                                      & (push only)\\
                & $a = \bm{\alpha}\circ(\Tilde{s}-f_p)$& (force direction)\\
                & $\frac{\left|\bm{\alpha}^r\right|}{\left|\bm{\alpha}^{r-1}\right| + \epsilon} \le 2.9$ & (max two contacts)\\
                & $\left|s - s^* \right| \le tol$                                         & (path tolerance) \\
                & $s=s_0 + P\left(a - g\right)$              & (kinematics) \\
    \end{tabular}
    \egroup
\end{algorithm}

\makeatletter
\renewcommand*{\ALG@name}{Algorithm}
\makeatother
\setcounter{algorithm}{0}
\begin{algorithm}[tbp]
    \caption{Iteratively solve Problem \ref{alg:convex-fsp} with $n_{rw}$ iterations to more closely approximate Problem \ref{alg:nl-nc-fsp} with the cardinality constraint and more accurate approximation, $\Tilde{s}$.}
    \label{alg:reweighting}
        Initialize with $r = 0$, $\alpha^{r-1} = 0$\\
        \textbf{While} $r \le n_{rw}$ and $card(\alpha^r) > 2$:\\
        $\rightarrow$ Identify feasible set of footholds: $f \in \text{friction-cone}$\\
        $\rightarrow$ Solve Problem \ref{alg:convex-fsp}$(\alpha^{r-1})$\\
        $\rightarrow$ Update $\alpha^{r-1}$\\
        $\rightarrow r = r+1$
\end{algorithm}

\subsection{Cardinality through Reweighting Iterations}
The cardinality constraint is approximated by reweighting the costs on $\bm{\alpha}$ with an additional term:
\begin{align}
    f_{card} = \frac{\left|\bm{\alpha}^r\right|}{\left|\bm{\alpha}^{r-1}\right| + \epsilon}.
\end{align}
where $r$ is the reweighting iteration and $\epsilon$ prevents singularity. Equation \eqref{eqn:cardinality} is checked between reweighting iterations and considered a constraint for feasibility. This is constrained to be less than 2.9 such that there are no more than two contact points at any time, while allowing changes in the magnitude of the acceleration. This involves solving Problem \ref{alg:convex-fsp} iteratively, as per Algorithm \ref{alg:reweighting}, using the previous solution as a warm start.

\subsection{Phase/Side Assignment}
The phase/side assignment module takes the array of forces identified by the foothold selector and assigns each to a step to be taken by the left or right foot based on the contact position relative to the \ac{CoM} position. Consecutive forces from the same position are merged into single steps of a correspondingly longer duration. The foot trajectories are calculated using Hermite parametrizations here to indicate the assigned sides and phases.


\section{Results}\label{sec:results}
The resulting method can be applied to a wide range of environments and paths in real-time.

\begin{table*}[tbp]
    \centering
    \caption{Solve times (s) and required reweighting iterations for each environment, varying the horizon duration, and number of surfaces at each time step.}
    \label{tab:timings}
    \begin{tabular}{l|lll|ll|l}
         \textbf{Environment}&  \multicolumn{3}{c|}{\textbf{Time Horizon (T)} (k = 20)}&  \multicolumn{2}{c|}{\textbf{Surfaces at each Time (k)} (T = 1.5~s)}& \textbf{RW. Iterations}\\
         &  1.5~s (N = 10) &  3.0~s (N = 20) &  4.5~s (N = 30)&  k=10& k=20&\textbf{Avg./Min/Max}\\ \hline
        Flat Ground&$0.0442 \pm 0.0018$ & $0.154 \pm 0.004$ & $0.607 \pm 0.18$ & $0.0182 \pm 0.0042$ & $0.0417 \pm 0.013$ & $1.12/1/3$\\
Step Stones &$0.0357 \pm 0.0013$ & $0.183 \pm 0.0042$ & $0.958 \pm 0.25$ & $0.0178 \pm 0.0021$ & $0.0297 \pm 0.0012$ & $1.62/1/4$\\
Chasm&$0.0565 \pm 0.034$ & $0.17 \pm 0.05$ & $0.819 \pm 0.086$ & $0.0154 \pm 0.00044$ & $0.0431 \pm 0.0087$ & $1.38/1/6$\\
Staircase (Up)&$0.169 \pm 0.031$ & $0.334 \pm 0.26$ & $0.592 \pm 0.012$ & $0.016 \pm 0.00052$ & $0.0407 \pm 0.01$ & $1.34/1/10$\\
    \end{tabular}
\end{table*}
\begin{figure*}[tbp]
    \centering
    \includegraphics[width=\linewidth]{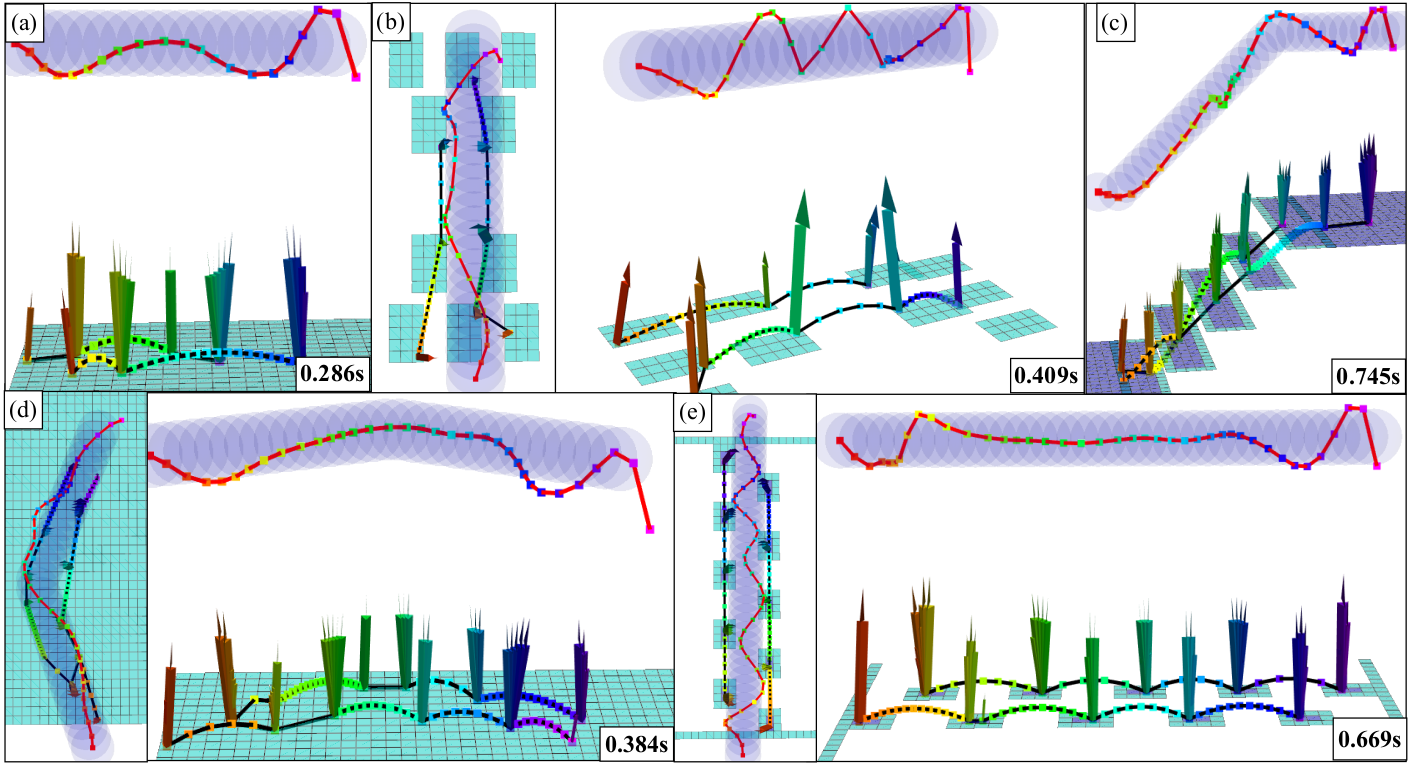}
    \caption{Trials in a range of environments with different desired trajectories. The maximum CoM path deviation is constrained to 0.1~m (transparent blue circles), and the CoM path and force arrow colors relate to time. Two arrows of the same color indicate a double support phase. (a) 3s horizon, straight path on flat ground; (b) 3.4s horizon, step stones of varying heights with a gap; (c) 5.1s horizon, climbing up a staircase; (d) 5.1s horizon, bend in the path; (e) 6.1s horizon, discrete step stones. We found that the more challenging environments (b, c, and e) required 2-3 reweighting iterations to meet the cardinality constraint.}
    \label{fig:results}
\end{figure*}

\subsection{Fast Solver}
\begin{figure}[tbp]
    \centering
    \includegraphics[width=1\linewidth]{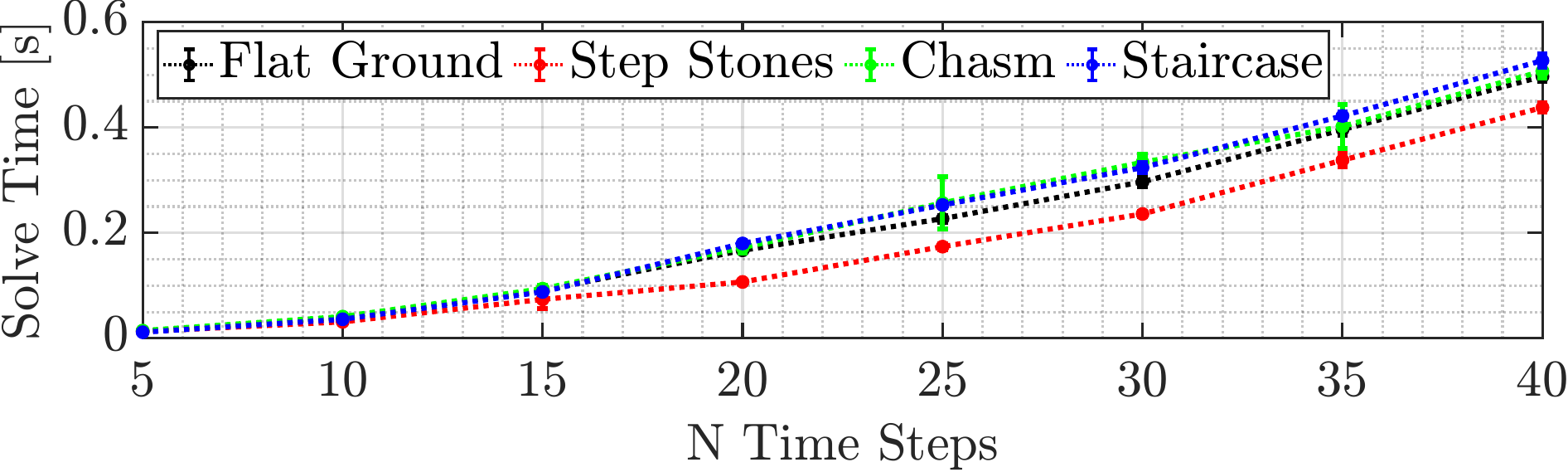}
    \caption{Trend in solve time with increasing number of time steps for the flat ground environment with k=20, dt=0.15~s. On a log-log plot the slope is 1.7, indicating almost quadratic growth in computation time with planning time horizon. Each trial required one reweighting iteration.}
    \label{fig:time-graph}
\end{figure}

This algorithm is 
solved with MOSEK \cite{mosek} on an Intel Core i7 CPU/3.00GHz 8-Core desktop. A single iteration for a 1.5~s horizon is solved in under 40~ms. Adding reweighting iterations increases the solve time almost quadratically (after the first iteration). The solver converges to a minimal cardinality within six iterations in most cases, with the solve time still significantly less than the time horizon as shown in Table \ref{tab:timings}. Each test was conducted with a desired forward velocity of 0.32~m/s, sample time of 0.15~s and 4 reweighting iterations, averaged over 10 trials. The impact of the number of time steps on the solve time is shown in Fig. \ref{fig:time-graph}.

Winkler et al. \cite{winkler_gait_2018} generate a center of mass motion and footstep plan with automatic gait discovery with a model of similar fidelity. The approach includes the centroidal angular momentum which we approximately constrain to zero (as per Equation \eqref{eqn:force-dir-constraint}). Testing the code provided by Winkler on the same computer with a 1~m path on flat ground and the same configuration and corresponding parameters (robot height 0.5~m, 20 time steps), a solution to the program presented in Winkler was found in 1.77~s, while a solution to Problem \ref{alg:convex-fsp} was found in 0.142~s (requiring one reweighting iteration).

\subsection{Environment Versatility}

The results for a subset of trials are shown in Fig. \ref{fig:results}, including flat ground, a bent path, discrete step stones, a chasm, and a staircase.
With each surface point in the environment considered separately, there is no assumption of consistent terrain characteristics allowing the algorithm to plan over terrain with inconsistent features.



\subsection{Automatic Gait Discovery}
With the intended application to bipedal robots, the cardinality is constrained to having no more than two contact points at any given time. While this does allow for consecutive double support phases, this has not been observed to occur. The reweighting iterations cause the robot to favor single support phases which are observed to alternate sides automatically, despite this not being imposed in the problem. This allows for some cases where the robot hops or jumps off two feet which can be desirable in more challenging environments. Additionally, the number of steps in the trajectory and the durations of the phases are unconstrained, providing more flexibility in the problem.

The ideal walking trajectory is observed to occur when the desired velocity is around 0.32~m/s for a robot with the torso at a height of 1~m. This corresponds to the resonant frequency of a pendulum with a length of 1~m. This is observed on the flat ground environment and when ascending and descending the staircase, as shown in Fig. \ref{fig:results}. Testing slightly higher velocities leads to a succession of shorter steps, subject to limitations imposed by the local terrain.

\subsection{Tuning the Objective Function}
The weights in the cost function determine the nature of the gait cycle. For instance, balancing the $l_2$-norm in the temporal consistency term with the $l_1$-norm in the cardinality approximation determines the sparsity and smoothness of the forces, favoring single or double support phases respectively. Similarly, the weights on the penalty on jerk and the path deviation cost balance the smoothness and accuracy of the \ac{CoM} trajectory.

\subsection{Stability and Feasibility}
As observed in Fig. \ref{fig:results}, while periodic motion is achieved and there is a penalty the jerk of the \ac{CoM}, the behavior at the end of the path needs further exploration to ensure stability in a receding horizon framework. However, the solve times would enable the planner to run at a high frequency for short horizons (10~Hz), which has a strong impact on the robustness \cite{apgar2018fast} and the ability to prevent falls with steps at capture points. While there is the approximation of the \ac{CoM} position for the SLIP leg vectors, this algorithm could provide the input to algorithms that use more accurate robot models but require a contact schedule \cite{apgar2018fast}.



\section{Conclusion}
We generate coupled center of mass motion and footstep plans that follow provided headings while accounting for traversability constraints substantially faster than real-time. This provides a contact schedule and \ac{CoM} trajectory approximation for a SLIP-like bipedal robot model. This method is a new representation of the bipedal footstep planning problem to be explored further.

\bibliographystyle{IEEEtran}
\bibliography{IEEEabrv,references}

\end{document}